# Agricultural Object Detection with You Look Only Once (YOLO) Algorithm: A Bibliometric and Systematic Literature Review

*Chetan M Badgujar[a], Alwin Poulose[b] and Hao Gan[a]*

[a]Biosystems Engineering and Soil Sciences, The University of Tennessee, Knoxville, TN 37996.

[a]School of Data Science, Indian Institute of Science Education and Research Thiruvananthapuram (IISER TVM), Vithura, Thiruvananthapuram, 695551, Kerala, India.

**Abstract:**

Vision is a major component in several digital technologies and tools used in agriculture. Object detection plays a pivotal role in digital farming by automating the task of detecting, identifying, and localization of various objects in large-scale agrarian landscapes. The single-stage detection algorithm, You Look Only Once (YOLO), has gained popularity in agriculture in a relatively short span due to its state-of-the-art performance in terms of accuracy, speed, and network size. YOLO offers real-time detection performance with good accuracy and is implemented in various agricultural tasks, including monitoring, surveillance, sensing, automation, and robotics operations. The research and application of YOLO in agriculture are accelerating at a tremendous speed but are fragmented and multidisciplinary in nature. Moreover, the performance characteristics (i.e., accuracy, speed, computation) of the object detector influence the rate of technology implementation and adoption in agriculture. Therefore, the study aims to collect extensive literature to document and critically evaluate the advances and application of YOLO for agricultural object recognition tasks. First, we conducted a bibliometric review of 257 selected articles to understand the scholarly landscape (i.e., research trends, evolution, global hotspots, and gaps) of YOLO in the broad agricultural domain. Secondly, we conducted a systematic literature review on 30 selected articles to identify current knowledge, critical gaps, and modifications in YOLO for specific agricultural tasks. The study critically assesses and summarizes the information on YOLO's end-to-end learning approach, including data acquisition, processing, network modification, integration, and deployment. We also discussed task-specific YOLO algorithm modification and integration to meet the agricultural object or environment-specific challenges. In general, YOLO-integrated digital tools and technologies show the potential for real-time, automated monitoring, surveillance, and object handling to reduce labor, production cost, and environmental impact while maximizing resource efficiency. The study provides detailed documentation and significantly advances the existing knowledge on applying YOLO in agriculture, which can greatly benefit the scientific community. The results of this study open the door for implementing YOLO-based solutions in practical agricultural scenarios and add to the expanding corpus of information on computer vision applications in agriculture.

## Introduction:

In an era of digital agriculture, the application and integration of advanced sensing, data processing, analytical, and digital tools are becoming more accessible (Hamidisepehr et al., 2020; World Bank, 2019). Digital agricultural tools and technologies can help optimize or gain resource efficiencies to increase food production with minimal environmental footprints (Fountas et al., 2020; Schroeder et al., 2021). Therefore, agricultural digitization and automation have been perceived as one of the key solutions to the grand challenges of global food systems, including increased production costs, labor shortages, declining or degrading natural resources, climate change, etc. (McFadden et al., 2023).

Continuous advancement in visual sensor technology, computational capabilities, and data-driven machine learning methodologies has led to a growing interest in digital agricultural technologies, mainly automated and intelligent systems. The deep learning-based computer vision applications, including object recognition methods, have exponentially increased in agriculture, primarily due to reduced hardware costs (e.g., camera, storage, and computational system) and increased computational capabilities in recent years (Liu et al., 2023; Tian et al., 2023). Object recognition mimics human perception systems with an integration of cameras, image processing methods, and deep learning models to detect, identify, and track objects of interest within images/video. Agriculture involves dealing with multiple objects; hence, object detection has found numerous applications in crop and animal farming, including aquaculture. This application extends well beyond remote sensing or monitoring (Zhang et al., 2023), automated or robotic systems (An et al., 2022; Liu et al., 2022; Xu et al., 2022), decision support tools, and more. Object detection is a key component of digital agriculture and can advance precision farming practices by reducing labor and costs while optimizing resources, ultimately increasing agricultural productivity and production.

Deep learning employs convolutional neural network (CNN) architectures and is at the forefront of object detection models. Deep learning techniques facilitate automatic feature learning and representation from large labeled image datasets, achieving state-of-the-art performance (i.e., accuracy and speed) compared to conventional image processing methods that require manual feature extraction (Goodfellow et al., 2016; LeCun et al., 2015). Deep learning is rapidly evolving, and mainstream deep learning-based object detection algorithms include: (a) One-stage detectors such as Single Shot Multibox Detector- SSD (Liu et al., 2016), You Only Look Once- YOLO (Redmon et al., 2016), RetinaNet (Lin et al., 2018), CenterNet (Duan et al., 2019) and EfficientDet (Tan et al., 2020). (b) Two-stage detectors such as Region-based networks that are RCNN (Girshick et al., 2014) and RCNN family including Fast RCNN (Girshick, 2015), Faster RCNN (Ren et al., 2016), Mask RCNN (He et al., 2018), Cascade R-CNN (Cai & Vasconcelos, 2017). (c) Vision transformers such as Detection Transformer- DETR (Carion et al., 2020) and DETR series (Shehzadi et al., 2023). These algorithms differ in design principles and architecture, offering speed, accuracy, efficiency, and algorithm size trade-offs. For example, a two-stage detector has better accuracy but slower speed, while a one-stage detector balances speed and accuracy with relatively higher speed (Zhang et al., 2022). However, DETR models, on the other hand, provide end-to-end object detection and simplify the overall architecture but are highly computational and large, making them unsuitable for embedded devices (Shehzadi et al., 2023). The choice of algorithm depends on the task being undertaken, but higher speed (near real-time), good accuracy, and smaller model size are usually desirable for agricultural tasks. Hence, the one-stage detector, YOLO, has gained popularity in agriculture in a relatively short span due to its suitability for real-time capabilities with good accuracy and compatibility with resource-constrained devices.

Vision is a major agricultural component in several digital technologies and tools for monitoring, surveillance, sensing, automation, and robotics operations. Agricultural objects vary on spatial and temporal scales, often in complex and unstructured environments (Bechar & Vigneault, 2016). Thus, a generic object detection algorithm trained for agricultural tasks either fails or delivers poor performance. Nevertheless, the performance characteristics (i.e., accuracy, speed, computation) of the object recognition model influence the implementation and adoption of technology. For example, a vision-based fruit-picking robot's detection accuracy determines the yield, and its speed determines the harvesting throughput. However, both research and application of YOLO in agriculture are accelerating at a tremendous speed but and are often fragmented and non-interdisciplinary, which needs to be documented and critically assessed. Therefore, the study aims to collect extensive literature to document and critically evaluate the advances and application of YOLO for agricultural object recognition tasks. The comprehensive article collection would be critical to understanding the scientific landscape of YOLO applications in agriculture. The study would also explore the end-to-end YOLO model development pipeline and major modifications or improvements adopted to improve the model performance for specific agricultural tasks. Moreover, the paper discusses the YOLO integration with state-of-the-art algorithms and its strengths and limitations for object detection tasks. The rest of the paper is organized as follows. Section 2 presents the Yolo overview and Section 3 discusses the literature review methodology in agricultural object detection with the YOLO algorithm. In Section 4, we explained the results and discussion and Section 5 concludes our paper with future research directions.

## 2. YOLO Overview

YOLO, being a single-stage detector, performs both localization and classification tasks in a single pass through the network architecture. A working principle of the YOLO algorithm is briefly explained in Figure 1(a), which involves the following steps: (a) Grid division: an input image is divided into a fixed-size grid (S × S), where each cell is responsible for making predictions on object localization (bounding box) and classification (class probabilities). (b) Bounding box and class prediction: each cell predicts multiple bounding boxes, along with box location (x, y), dimensions (w, h), and confidence score. Also, class probabilities are predicted for each bounding box, which refers to confidence that the detected objects belong to a predefined class. (c) Final detection: the algorithm computes a final score for each bounding box, considering the box confidence score and class probabilities. This results in duplicate overlapping boxes with high and low-confidence detections. Hence, a non-maximum suppression (NMS) algorithm is applied to retain the bounding box with the highest score and remove overlapping or low-confidence detections. The final prediction includes a bounding box (x, y, w, h), class label, and confidence score.

The primary architecture of YOLO includes three major components (Figure 1b), which are described as follows: (1) Backbone contains a series of layers (convolutional, pooling) and blocks (C3, SPPF, C2F). The backbone down-samples the input image spatially while increasing the feature maps. It captures the hierarchical image features at different scales and resolutions, serving as a feature extractor where initial layers extract low-level and deeper layers extract high-level features. (2) Neck forms a connection between the backbone and head with a series of intermediate layers, including convolutional, Spatial Pyramid Pooling (SPP), and Feature Pyramid Network (PANet) (Lin et al., 2017). It refines the features extracted by the backbone and enhances the spatial and semantic feature representation at different scales to handle objects of various sizes. (3) Head, contains multiple convolutional layers accompanied by activation functions to make final predictions (i.e., bounding box and class probabilities) based on features extracted by the backbone and neck. At the post-processing stage, it performs NMS operations to refine the final detections.

In 2015, Redmon et al., (2016) first introduced an end-to-end object detection framework called YOLO, inspired by a human vision system of looking only once to gather visual information. It achieved a state-of-the-art performance on generic object detection with real-time detection capabilities and good accuracy. Since then, the YOLO family has continuously evolved and gone through several versions, each introducing improvements in accuracy, speed, and network features. Currently, more than fifteen YOLO versions are available for generic object detection in an open-source domain, and the timeline of YOLO model development is presented in Figure 1c. Details on each YOLO version (Table 1), network modification, and contribution can be found in (Terven et al., 2023). The agricultural community has kept pace with the YOLO development, and it is widely adopted for agricultural object detection, with numerous evidence available in the literature. Unlike generic object detection, complex and unstructured agriculture environments pose diverse challenges. Therefore, many application studies reported minor or complete modifications in the YOLO network to suit their requirement better. This paper briefly discuss such (agricultural) domain-specific modifications in the YOLO network.

**(a) A working principle of the You Only Look Once (YOLO) algorithm for object detection tasks.**

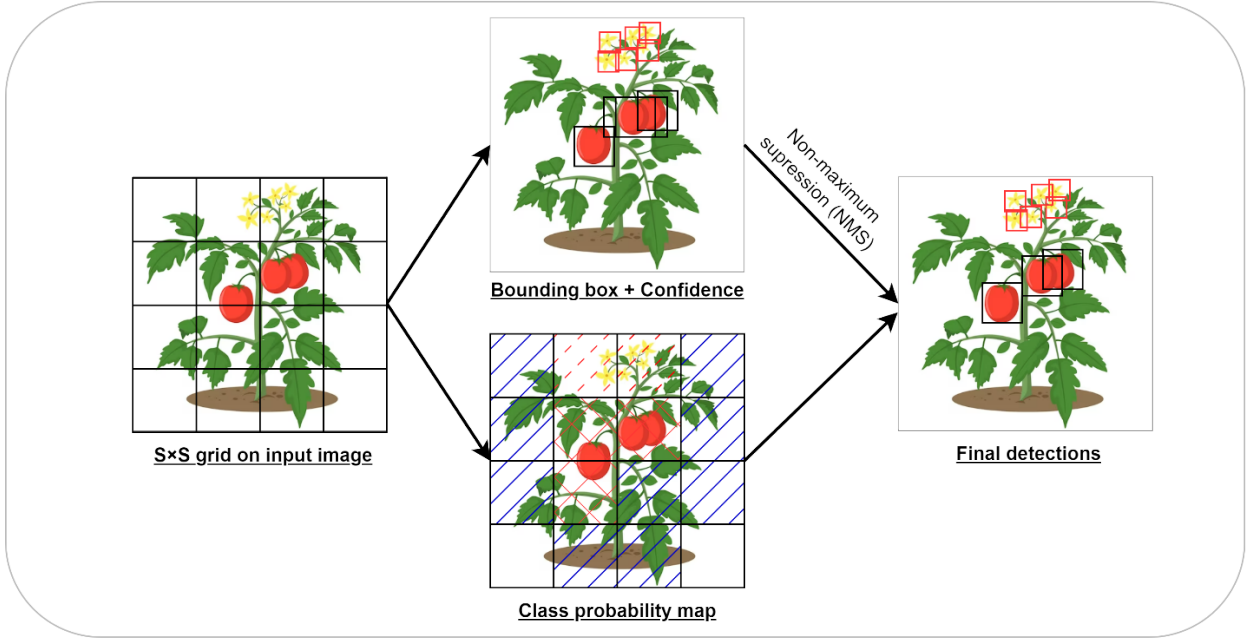

**(b) A primary network architecture of the You Only Look Once (YOLO) algorithm.**

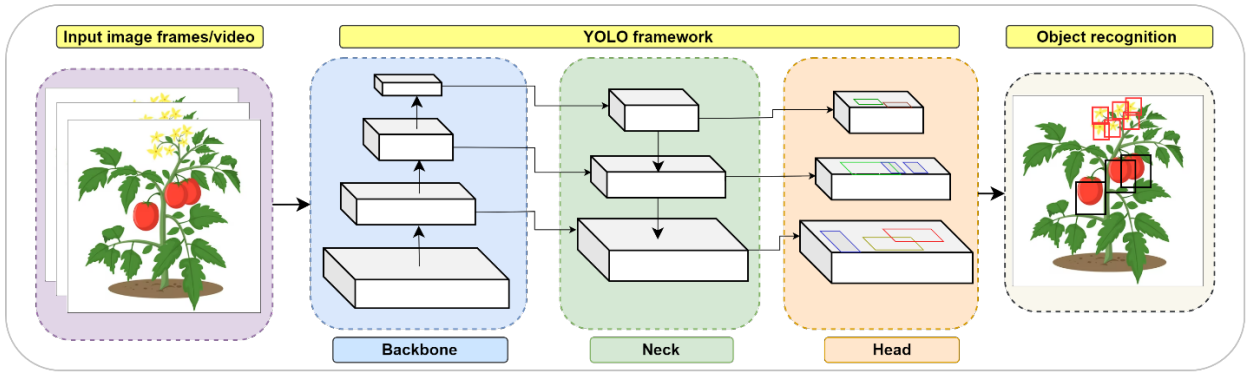

**(c) A brief timeline of the You Only Look Once (YOLO) algorithm development.**

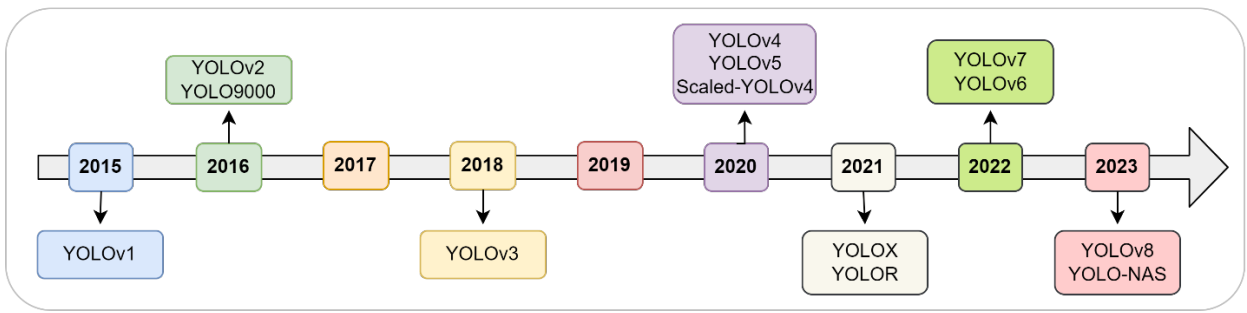

*Figure 1: The general overview of the You Only Look Once (YOLO) framework.*

Table 1: YOLO version and publicly available source code.

| YOLO version | Source code (Github repository) | Reference |
| --- | --- | --- |
| YOLOv1 | Not available | (Redmon et al., 2016) |
| YOLOv2 | https://pjreddie.com/darknet/yolov2/ | (Redmon & Farhadi, 2016) |
| YOLOv3 | https://pjreddie.com/darknet/yolo/ | (Redmon & Farhadi, 2018) |
| YOLOv4 | https://github.com/AlexeyAB/darknet | (Bochkovskiy et al., 2020) |
| Scaled-YOLOv4 | https://github.com/WongKinYiu/ScaledYOLOv4 | (Wang et al., 2021) |
| YOLOv5 | https://github.com/ultralytics/yolov5 | (Jocher, 2020) |
| YOLOX | https://github.com/Megvii-BaseDetection/YOLOX | (Ge et al., 2021) |
| YOLOR | https://github.com/WongKinYiu/yolor | (Wang et al., 2023) |
| YOLOv6 | https://github.com/meituan/YOLOv6 | (Li et al., 2022) |
| YOLOv7 | https://github.com/WongKinYiu/yolov7 | (Wang et al., 2022) |
| YOLOv8 | https://github.com/ultralytics/ultralytics | (Jocher et al., 2023) |
| YOLO-NAS | https://github.com/Deci-AI/super-gradients | (Aharon et al., 2021) |

## 3. Literature Review Methodology

The review process involves an extensive document search, collection, careful selection, systematic evaluation, and analysis of existing literature on YOLO application in the broad agricultural domain. Figure 2 provides a detailed procedure and steps followed during the review process.

### 3.1. Materials Collection

The relevant material collection is the primary step in the literature survey. Therefore, a keyword-based search was conducted on scientific databases to collect current and available research material. Several databases are available for material collection, including IEEE Xplore, Google Scholar, ScienceDirect, Web of Science, and Scopus. In this study, we chose the extensive and trustworthy Scopus database, which provides coverage to journals published globally. It includes non-English content, offers interdisciplinary field coverage, and encompasses peer-reviewed articles, non-journal content, books, conferences, and patents (Pellack, 2023; Rejeb et al., 2022). Other datasets were not explored to avoid material duplication unless the initial search results were minimal in number. The Scopus search included: (1) a keywords search, with a query: ["YOLO"] OR ["You Only Look Once"] AND ["Agriculture" OR "Farming"], (2) with a period filter of 2015-2024, since YOLO was first introduced in 2015. The search was conducted in November 2023, which resulted in 420 documents, including articles (e.g., research, review, and data), conference proceedings, and book chapters. The collected document dataset is large. Hence, a single review method will not help understand the trend and context.

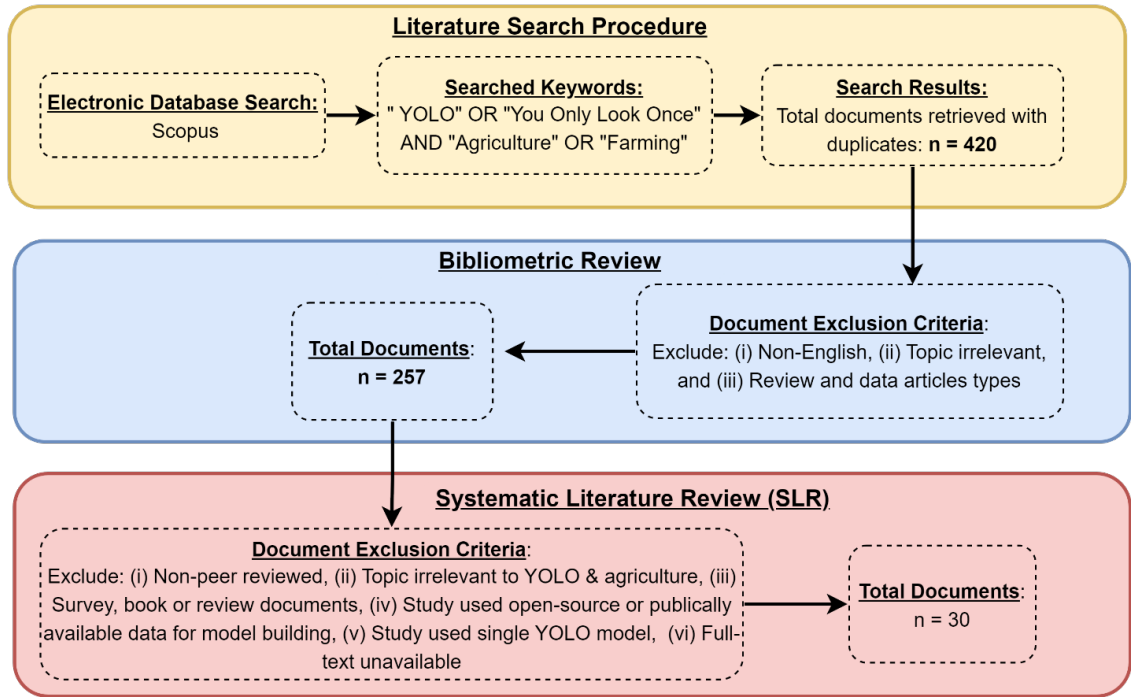

*Figure 2: An overview of an employed literature review procedure.*

### 3.2. Bibliometric Review

The bibliometric review summarizes extensive bibliometric data to provide an overview of the intellectual structure and emerging trends on a specific research topic (Rejeb et al., 2022). It is often used when a review scope is broad and the dataset is too large for manual review and inspection (Donthu et al., 2021). The bibliographic database offers a targeted information search on a specific topic and contains descriptive records of published scientific materials (e.g., articles, books, conference proceedings). It includes information on publications, authors, journals, and other bibliographic elements such as publication title, abstract, year, place, keywords, source, publisher, etc. The bibliographic data of 420 documents were exported from the Scopus database and stored in multiple bibliographic file formats (such as BibTex, RIS, XML, TXT, and CSV) for subsequent analysis. In an Excel CSV file, the bibliographic database (420 documents) underwent further sorting, cleaning, filtering, and elimination based on (1) language (excluding non-English documents), (2) topic relevance (agriculture and YOLO), and (3) article types (excluding review and data articles). This refinement process resulted in 257 documents, which is still too large for the manual review. Therefore, a bibliographic review was conducted on 257 documents to gain insight into the scientific landscape of YOLO application in agriculture, along with research status, trends, and evolution. However, the bibliographic review lacks the rigorous, impartial, and literature-wide assessment of individual research studies regarding quality and design. Therefore, a subsequent systematic literature review will follow a meticulous document selection procedure, record-keeping, review, evaluation, and assessment, aiming to critique the quality and limitations of the existing literature (Mengist et al., 2020). The goal is to identify current knowledge and critical gaps and needs, ultimately contributing to an improvement in the direction of future research.

### 3.3. Systematic Literature Review

Systematic literature review (SLR) involves systematic material collection, critical evaluation, detailed summary, and presentation on a topic of interest (Pati & Lorusso, 2018). An SLR summarizes and synthesizes the findings of existing research literature. It is used when a review scope is specific, and the dataset is small and manageable enough that its content can be manually reviewed (Budgen & Brereton, 2006; Donthu et al., 2021; Pati & Lorusso, 2018). Therefore, the following selection criteria were employed to limit the number of publications in SLR.

### 3.3.1. Document Selection Criteria

The bibliographic data (257 documents) was further scrutinized to select relevant documents based on the SLR method (Budgen & Brereton, 2006) that included title, abstract, and keyword screening. All documents were imported to a spreadsheet, and each document was verified against the predefined selection and exclusion criteria, which included a list of questions that set the boundaries for the literature review. The selection or exclusion questions are as follows:

Q1. Is a document related to YOLO and agriculture or farming?
Q2. Is a document a survey, book chapter, or review article?
Q3. Is a document full text available?
Q4. Is a document peer-reviewed?
Q5. Does a document employ open-source data for YOLO model building?
Q6. Does a document employ more than one object detection model?

The document selection criteria resulted in 30 documents, which were used for further detailed and systematic review. The review focused on the following criteria: (1) application area, (2) image data source, (3) data processing techniques such as data augmentation, (4) YOLO versions used and adopted YOLO improvements or modification, and (5) YOLO integration with other algorithms.

## 4. Results and Discussion
### 4.1. Bibliographic Analysis

Bibliometrics is a statistical analysis of bibliographic data that contains unstructured text information (DeGroote, 2023). It includes a rich source of text and keyword associations with published documents on YOLO and agriculture in scientific literature. The bibliographic analysis offers: (1) quantitative (evaluation and interpretation) and qualitative (interpretation) analysis (Donthu et al., 2021). (2) insights into the scholarly landscape, such as research trends, evolution, global hotspots, emerging research topics, gaps, and its adoption or integration status (DeGroote, 2023; Rejeb et al., 2022).

### 4.1.1. Research Trends and Evolution

A detailed summary of published documents on the YOLO application in agriculture is presented in Figure 3. The document collection was organized based on publication frequency on a time scale (year) to understand the overall traction and trend. YOLO was first introduced in 2015, and in a short period, it has gained increasing attention in the agricultural community, with an exponential increase in documents (Figure 3a) from 2018 to the present (November 2023). This highlights the necessity and significance of the proposed review to cover the most prominent and relevant work. The number of documents was relatively low until 2019, with a significant increase observed in subsequent years. This increase might be attributed to (1) rapid advancements and subsequent release of a novel YOLO version with improvements in speed and accuracy, (2) state-of-the-art performance benchmarks in terms of real-time inference with good accuracy, and (3) contributions from original authors and the open-source community. The dataset contained documents published until November 2023; hence, documents from 2023 are limited, but they indicate an ongoing popularity and research interest in the field.

Next, the document collection was organized based on publishers and document types. The number of documents published by top publishers is presented in Figure 3b, which shows the commonly preferred platforms for disseminating agricultural research. It also highlights that the Scopus database contains documents from multiple publishers. The most common document types are research and conference papers (Figure 3c). Figure 4 illustrates the geographic distribution of affiliated authors in a document collection to understand the research hotspots and individual nations' contributions. Many documents originate from China, India, the US, and Brazil, which are prominent agricultural production leaders across several crop categories.

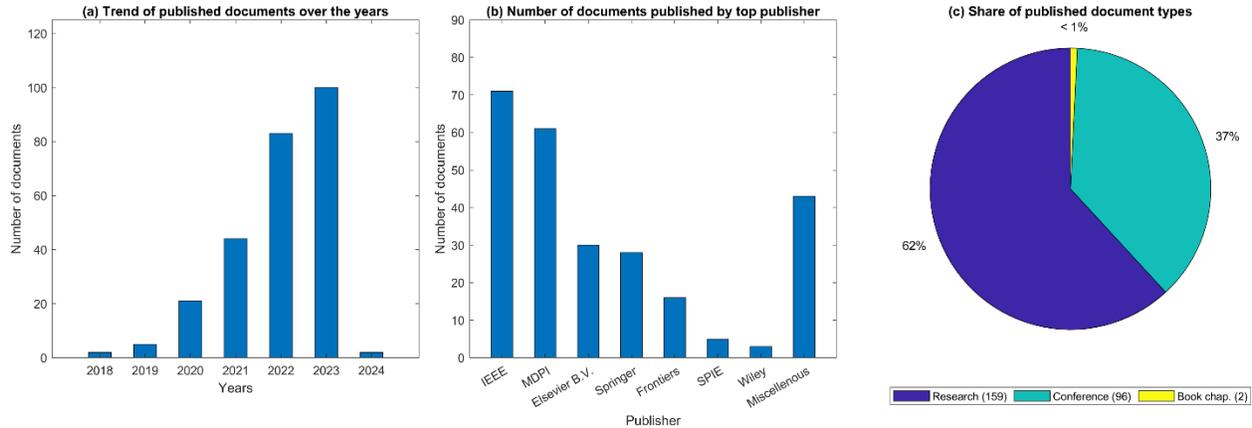

*Figure 3: A summary of published documents (n = 257): (a) trend of published documents over the years, (b) number of documents published by top publishers, and (c) share of published document types.*

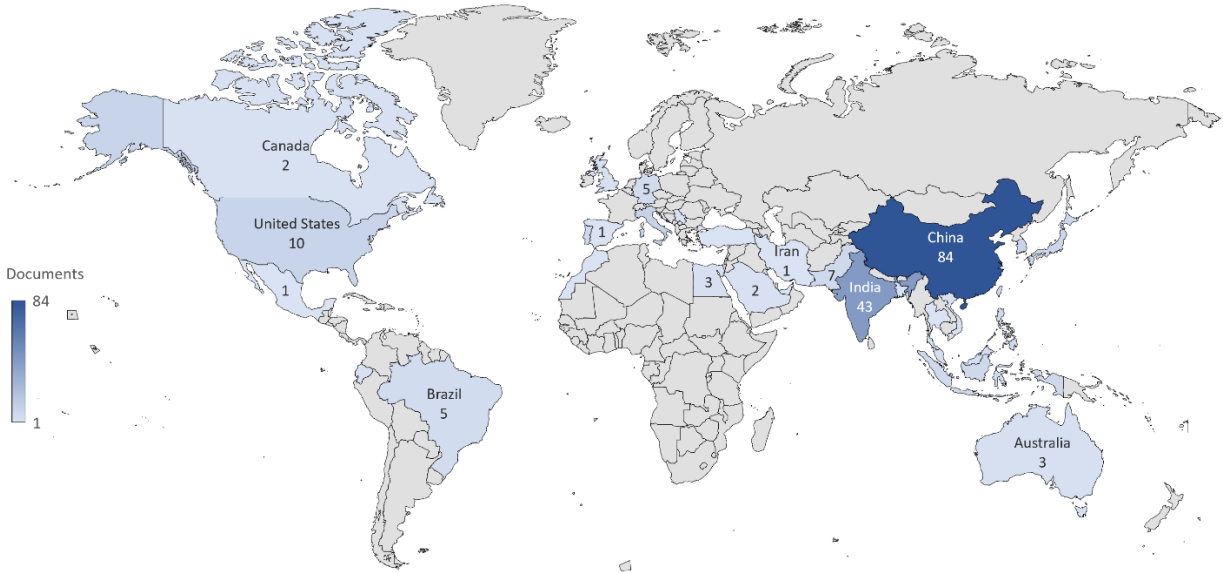

*Figure 4: Global distribution of affiliated authors in the analyzed documents.*

### 4.1.2. Keyword Analysis

Keywords can effectively describe the content of published literature. The keyword co-occurrence in a document suggests a semantic relationship, and the higher co-occurrence frequency implies a greater correlation. In bibliometrics, keyword analysis involves the collection, semantic mapping, visualization, and analysis of the keywords associated with scientific documents on a specific topic of interest (Aria & Cuccurullo, 2017). Keyword analysis can help explore a current research status, major focus areas, and emerging or potential future topics from literature documents (Chen & Xiao, 2016; Pesta et al., 2018; Wang & Chai, 2018). It includes co-occurrence, frequency, and network analysis, using keywords to create a visual map of a knowledge domain. The visual map illustrates a semantic relationship among several keyword topics and their co-occurrence frequency from literature keyword data. In this study, we conducted a keyword analysis on both author-supplied and index keywords using a software tool called "VOSviewer." The VOSviewer employs text mining and natural language processing (NLP) techniques to

generate keyword networks (Bukar et al., 2023; van Eck & Waltman, 2010). The co-occurrence network (Figure 5 & 6) was developed on author-supplied and index keywords. The keywords from 257 documents were used for the analysis, and a minimum of five keyword co-occurrences were selected. We also eliminated the semantically identical keywords such as "machine vision" and "computer vision" or "drones" and "UAV" and so on. The generated co-occurrence network/map is shown in Figure 5 and 6, which contains three integral elements described as follows: (a) Node/circle: represents individual keywords, and a node or label size determines the keyword frequency. Therefore, a larger node or label represents more frequently co-occurring keywords. (b) Link/line: represents the strength of co-occurrence; thicker lines indicate a strong, while thinner lines indicate a weak relationship. (c) Layout: the nodes are spatially arranged on a map to show the relatedness of keywords based on co-occurrence. The shorter distance between two keywords shows stronger relatedness.

In scholarly literature, author keywords are the terms created or selected by authors and considered a core element to summarize and convey the scientific content (Kwon, 2018; Lu et al., 2020). The co-occurrence network of author keywords is shown in Figure 5, which highlights a growing prominence of terms such as "Deep learning," "object detection," "YOLO," "Convolutional Neural Network (CNN)," "Computer vision," and "precision agriculture." These keywords are centrally positioned with strong interconnected links, underscoring their significant association. The popular object detection models were, namely, "YOLOv3", "YOLOv4", "YOLOv5", "YOLOv7", and "Faster R-CNN," which are placed along the edge with low to moderate link strength, suggesting low occurrence and relatedness, except "YOLOv5". This might be because the authors often preferred the generic keyword "YOLO" and refrained from using specific YOLO versions. Notably, "YOLOv5" stands out with the highest occurrence and relatively stronger link strength (Fig. 5). The keywords associated with agricultural objects or tasks included "pest detection," "disease detection," and "weed detection." These keywords are positioned at the network's periphery, indicating limited availability of published research and potential gaps or unexplored areas. The keyword "object detection" may include agricultural objects such as fruits, vegetables, or other objects, further explained in a detailed Sankey diagram (Fig. 7).

The co-occurrence network also included keywords like "smart agriculture," "precision agriculture," and "precision livestock farming." These keywords are interconnected concepts in modern agricultural practices, emphasizing the integration of sensors, data, models, and advanced technologies to enhance production efficiency and facilitate informed decision-making on input applications (Shaikh et al., 2022; Sharma et al., 2022). Mainly, "smart agriculture" and "precision agriculture" hold central positions in the network with the highest co-occurrence and link strength, indicating emerging research topics. Conversely, the keyword "precision livestock farming" is situated at the network's periphery, meaning a potentially unexplored research area and a potential research gap from collected documents. The keywords such as "Internet of Things (IoT)" and "Unmanned Aerial Vehicle (UAV)" are associated with advanced technologies and tools with increased adoption and utilization in modern agriculture.

The indexed keywords are the terms or phrases assigned to scholarly documents from the dataset thesaurus, which describe a specific concept, idea, or covered subject (Kraus 2023; Vukic 2023). The co-occurrence network of 48 indexed keywords is shown in Figure 6. Like author keywords, the primary and frequent keywords are placed in the center, and a few important new keyword clusters are discussed as follows. Apart from "weed," "disease," and "object detection," the indexed keyword map includes additional keywords such as "animal," "plants," "fruits," "orchards," "aquaculture-fish," "crops," "cotton," "forestry," "farms," "harvesting," and "cultivation." These keywords are distributed on the map, and their occurrence frequency and link strength can be inferred from Figure 6. The presence and associations of these keywords on the map provide insights into current research trends and focus areas. For instance, the keyword "fruit" is interconnected with "orchards," "disease detection," "harvesting," "agricultural robots," "real-time," "precision agriculture," and many other keywords. This interconnection provides a comprehensive overview and qualitative interpretation of available research literature on fruits. By examining such links, one can infer the research evidence related to fruit harvesting or disease detection systems with real-time agricultural robots. Similar information can be deduced from interconnections for "animal" and "cotton" categories. Surprisingly, no keyword was associated with pest detection, which indicates a potential research gap in collected documents.

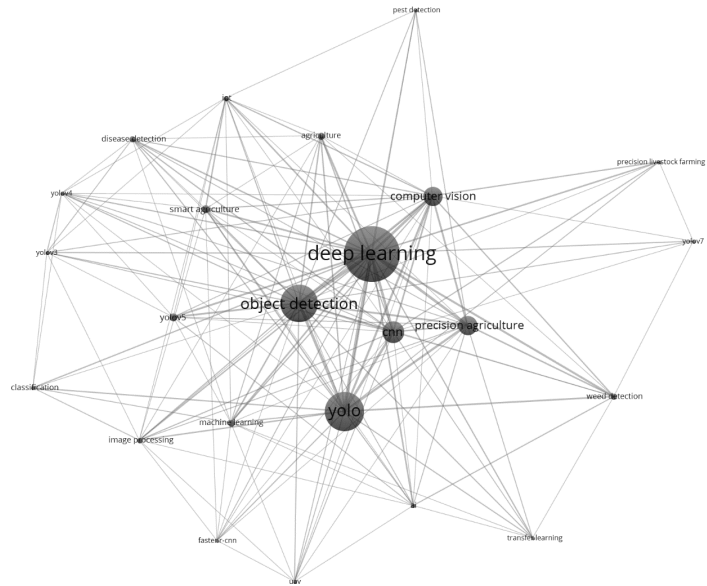

*Figure 5: Keyword analysis: Co-occurrence network of author-supplied keywords (total keywords = 634).*

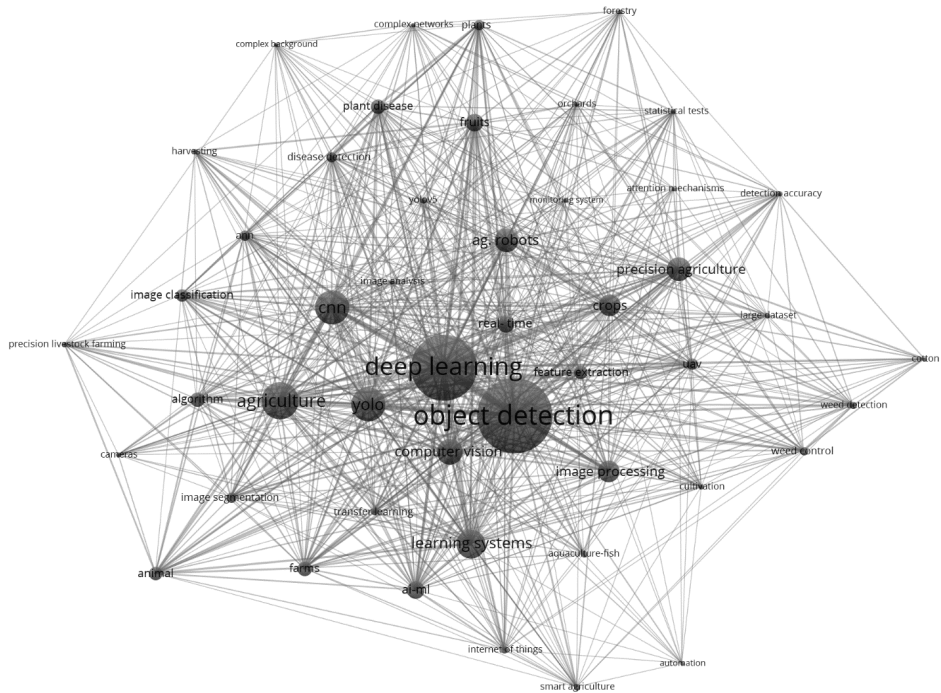

*Figure 6: Keyword analysis: Co-occurrence network of indexed keywords (total keywords = 1531).*

Next, the network incorporates keywords related to digital agricultural technologies and tools, namely, "real-time," "UAV," "Ag. robots," "automation," "IoT," "monitoring systems," "cameras," "algorithm," and others. These keywords are closely interconnected with "object detection" and "agriculture" since object detection is a core element of automation, robotics, and real-time monitoring systems performed with UAV-mounted cameras or IoT devices. Additionally, developing object detection models is a multi-step process that involves data collection, processing, model building, and evaluation. Therefore, the following keywords describe this process: "image processing," "feature extraction," "transfer learning," "image segmentation," "image analysis," "complex background," "large dataset," "attention mechanisms," "complex networks," and "detection accuracy." The network map generally provides quantitative information and interpretation to enhance understanding of the thematic structure and relationships in the YOLO application in agriculture.

A Sankey chart (Figure 7) illustrates an information flow for each document within the collected data, including multiple entities such as publication year, country, agricultural area, and employed YOLO model, along with its version. The aggregated information is a part of the SLR document selection process, inferred from the document title, abstract, and full-text skimming if required. The chart explains the diverse application areas within agriculture and the YOLO model version used in the collected data. This further highlighted a need for a systematic review and detailed information synthesis.

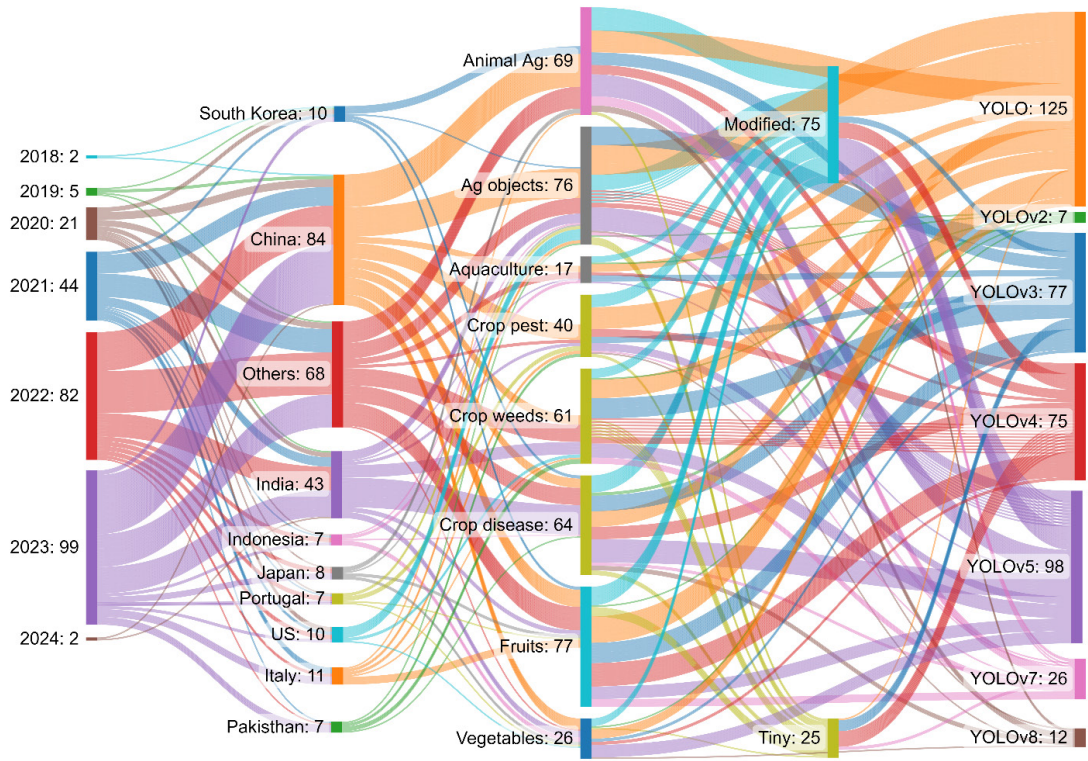

*Figure 7: Sankey chart showing an information flow for each document used in the study (n =257).*

### 4.2. Systematic Literature Review

YOLO is a unified and end-to-end learning approach where an input image is trained on a single neural network for object recognition. This approach directly predicts objects, eliminating intermediate steps that improve detection performance (i.e., speed, accuracy, and efficiency), and has contributed to the overall success of YOLO for real-time object recognition. A typical end-to-end learning pipeline used in YOLO is presented in Figure 8. The SLR aims to explore each YOLO end-to-end learning approach component and systematically review the collected literature to understand each aspect. The documents included in the SLR are presented in Table 2, summarizing information on the study goals, the model used, and the achieved experimental results. The included documents are from diverse agricultural domains, including crop production (commodity and specialty crops) and animal agriculture, including livestock and aquaculture. The best-performing model in each study is highlighted in bold letters, and model performance in terms of detection accuracy (mAP and average precision-AP) and speed (in frames per second-fps) is discussed for each study. Approximately 75% of the studies (23 articles) reported detection accuracy above 90%, and a few studies achieved a detection speed above 30 fps, which is considered real-time. This indicates the state-of-the-art performance characteristics of YOLO in agricultural object recognition. However, the detection performance was relatively lower, and accuracy ranged between 76% to 88% for small and dense objects such as coffee fruit (Bazame et al., 2021), apple flower (Chen et al., 2022), inflorescences (Xia et al., 2023), crop pests (Tian et al., 2023), tea buds (Li et al., 2023), and seedling maize weed (Liu et al., 2022). The SLR-selected studies implemented more than one model to understand where YOLO stands compared to other object detection models. Several studies compared YOLO's performance against SSD, Faster RCNN, RetinaNet, EfficientDet, and CenterNet. YOLO outperformed those models in either accuracy, speed, or model size, suggesting its superior performance for undertaken tasks. Besides, YOLO is available in several versions. However, many studies preferred modifying the YOLO algorithm, which will be discussed in detail.

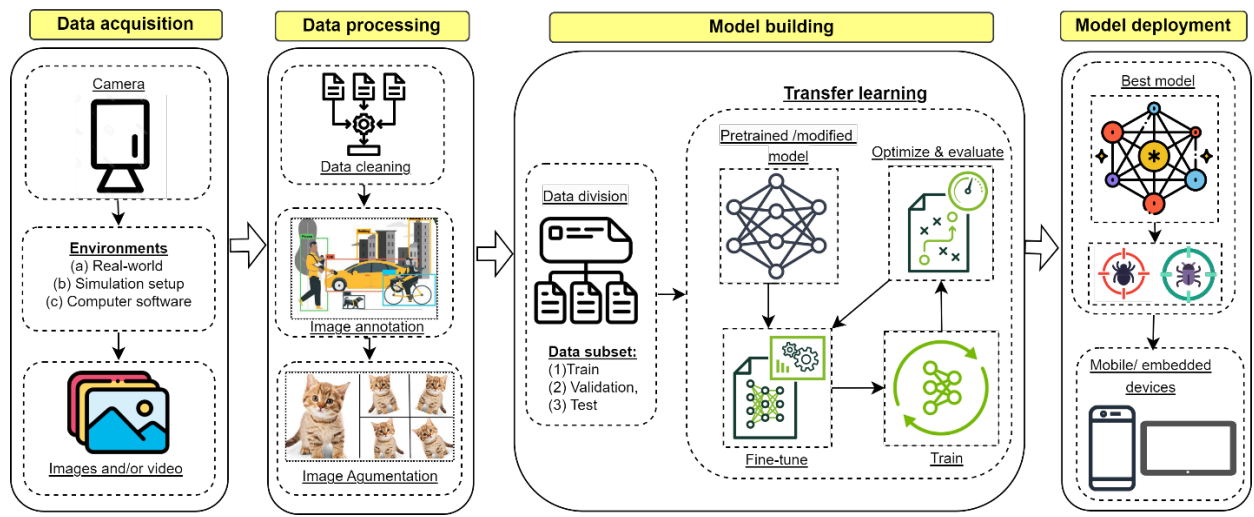

*Figure 8: An overview of the end-to-end learning approach followed in YOLO.*

### 4.2.1. Data Acquisition

The SLR-selected studies only included studies where data of the target objects was acquired from scratch. The data acquisition devices primarily had an RGB camera (e.g., high speed, action, industrial, surveillance, etc.) that is handheld (mobile phone, point-and-shoot), robot/drone mounted, or fixed positioned. The real-world object detection scenario affects the choice of acquisition devices, parameters, and approaches to capture diverse and high-quality images in an actual or simulated environment. The author sometimes preferred generating synthetic data with image processing techniques (Tian et al., 2023) or computer graphics software due to limited data generation abilities. A summary of data acquisition and processing methods employed in selected studies is presented in Table 3. The image resolution typically depends upon the camera hardware characteristics, object complexity, and detection task requirements. For example, the image resolution of 240×320 was selected to detect broiler chicken stunned state in slaughter lines (Ye et al., 2020), while to monitor large areas with drone a higher image resolution is preferred, provided individual object information is captured, as reported in studies detecting date palm trees (Jintasuttisak et al., 2022), maize tassels (Pu et al., 2023) and wheat spike (Zhao et al., 2022) with aerial imagery. The number of acquired images also varied greatly, but each study collected more than 100 images. There is no thumb rule to select the image data size for YOLO model building, and data size can vary based on task complexity, object diversity, and desired model performance. In general, the data size should be a good representative of the task undertaken, and fewer images with a large and diverse object can offer a good performance. For example, date palm trees data included 125 images with 8,797 objects, while fish uneaten pallet data included 175 images with 7,684 labels, both resulted in detection accuracy above 90% (Hu et al., 2021; Jintasuttisak et al., 2022). YOLO is a multi-class detection algorithm that achieved accuracy above 95% on a multi-class cotton weed detection task, including 12 weed classes (Dang et al., 2023).

### 4.2.2. Data processing

Data processing is a primary and multi-stage operation that aims to provide high-quality and relevant data for model development. The data cleaning process aims to eliminate images that do not contribute to the model learning for several reasons, including poor quality (noisy, low-resolution, blur, out-of-focus, outliers, occluded objects, etc.) and redundant (irrelevant, duplicate, etc.) images. In agriculture, data cleaning is often performed since image acquisition is often influenced by both machine (i.e., vibration, working speed, occlusion, etc.) (Bazame et al., 2021; Nasirahmadi et al., 2021) and environmental (i.e., lighting, weather, season, crop, or field type, etc.) factors. Next, a subject expert manually labels each image by drawing a rectangular bounding box and assigning class labels to each object. Presently, several online and offline data annotation tools and platforms are available for manual labeling.

Data augmentation is an optional image processing method that applies various image transforms to the original training images to create new training samples. The augmentation aims to artificially increase data size, quality, and diversity to ensure good model performance (Badgujar et al., 2023; Mumuni & Mumuni, 2022). The data augmentation transforms implemented in selected studies are listed in Table 3. These transforms can be grouped into: (1) Geometric transforms (e.g., flip, crop, rotate, translation, noise, pan, blur, scale), which introduce a variation in the spatial arrangement of images, simulating a real-world image with varying object pose, size, orientation, movement, and zoom. It makes models more robust to changes resulting from image acquisition systems. For example, image blur resembles camera movement or vibration. (2) Color space transforms (e.g., contrast, brightness, sharpening, histogram equalization, hue, saturation, shadow, sun flare, etc.) introduce the environmental variation, such as illumination in the images. (3) Image mixing transforms (e.g., mixup, cutout, mosaic) introduces a variation in image appearance, context, and composition to simulate complex scenes and backgrounds. It provides diverse images to improve model generalization. These transformations were applied randomly to increase the data size and diversity by multiplying the acquired images (Table 3). A large number of images can be produced using augmentation techniques, but the multiplication factor in selected studies ranged from 2 to 100. The data augmentation method improves the model performance in accuracy and robustness. Firstly, the detection accuracy was significantly improved for boiler chicken detection on slaughter line (Ye et al., 2020), fish behavior detection (Hu et al., 2021; X. Hu et al., 2021), pest detection in trap images (Tian et al., 2023), multi-class cotton weed detection (Dang et al., 2023) and poultry chickens detection (Chen et al., 2023). Secondly, augmentation can help build accurate and robust models. For example, (Mirhaji et al., 2021) used various transforms such as contrast, shadow, and sun flare to introduce illumination in orange orchard images. The trained models showed no significant difference between night and day, suggesting robustness. Data augmentation in YOLO training contributes to the model's generalization, robustness, and ability to handle diverse scenarios to ensure object detection models perform well on a wide range of real-world inputs. Data augmentation saves time and resources in manual labeling but is beneficial when obtaining sufficiently large and real-world data is impractical.

*Table 2: A selected document in an SLR process on YOLO application in agricultural object recognition*

| # | Study focus area | Implemented models | Performance |
|---|---|---|---|
| 1 | Stunned broiler detection on slaughter line (Ye et al., 2020) | YOLO, **YOLO+MRM*** | Accuracy: 96.1% |
| 2 | Sugar beet mechanical damage detection in harvester (Nasirahmadi et al., 2021) | **YOLOv4,** Faster RCNN (Inceptionv2 and NAS), RFCN-ResNet101 | Accuracy: ~90% and speed: 29fps |
| 3 | Orange detection (Mirhaji et al., 2021) | YOLOv2, YOLOv3, **YOLOv4** | Accuracy: 91.4% |
| 4 | Uneaten pellet detection in fish cage (X. Hu et al., 2021) | **Imp-YOLOv4,** YOLOv4, YOLOv3 | Accuracy: 92.6% |
| 5 | Coffee detection in harvester (Bazame et al., 2021) | YOLOv3-tiny (input image size: 416, 608, 704, **800**, 896) | Accuracy: 84% |
| 6 | Fish behavior detection and monitoring (Hu et al., 2021) | **Imp-YOLOv3-Lite**; Faster-RCNN, YOLO, YOLOv2, YOLOv3, SSD | AP: 89.7% and speed: 240fps |
| 7 | Real-time apple detection (Chen et al., 2021) | YOLOv4, **Des-YOLOv4**, Faster RCNN | Accuracy: 93.1% and speed: 51fps |
| 8 | Apple flower detection (Chen et al., 2022) | **Imp-YOLOv5s**, YOLOv5s, Faster RCNN, SSD, YOLOv3 | Accuracy: 77.5% |
| 9 | Wheat spikes detection (Zhao et al., 2022) | **OSWSDet (YOLOv5),** Faster RCNN, SSD, YOLOv5 | Accuracy: 90.5% |
| 10 | Date palm tree detection (Jintasuttisak et al., 2022) | SSD300, YOLOv3, YOLOv4, **YOLOv5** (s,l,**m**,x) | Accuracy: 92.3% |
| 11 | Strawberry growth detection (An et al., 2022) | YOLOv4, YOLOv5, YOLOX, **SDNet (YOLOX)** | Accuracy: 94.3% |
| 12 | Tomato detection and counting (Ge et al., 2022), | YOLOv5 (s,m,l), **YOLO-DeepSort**, YOLOv3 | Accuracy: 95.8% |
| 13 | Apple detection in orchard (Zhang et al., 2022) | **Imp-YOLOv4**, YOLOX, YOLOv3-efficientnetB0 | Accuracy: 95.7% |
| 14 | Seedling maize weed detection (Liu et al., 2022) | Faster RCNN, SSD300, YOLO-(v3, v4, v3-tiny, v4-tiny), **YOLOv4-Weeds** | Accuracy: 86.7% and speed: 57fps |
| 15 | Flower detection (Park & Park, 2023) | YOLOv3, YOLOv4-tiny, **YOLOv4-Circle** | Accuracy: 92.6% |
| 16 | Crop row (rice) detection in drone imagery (Ruan et al., 2023) | **Imp-YOLOR**, YOLOv4, Faster RCNN, SSD | AP: 91.8% and speed: 31.1fps |
| 17 | Cow behavior recognition and tracking (Zheng & Qin, 2023) | SSD, RetinaNet, EfficientDet-D1, YOLO-(v7, v8m, v5m), **PrunedYOLO-Tracker** | Accuracy: 88.2% and speed: 81fps |
| 18 | Grape stem detection (Wu et al., 2023) | **YOLOv5** (n,m,x), YOLOv3, SSD, Faster RCNN, CenterNet2 | Accuracy: above 90% |
| 19 | Tea buds detection (Li et al., 2023) | Faster RCNN, YOLO-(v3, v4, v4-tiny, v5s), SSD, EfficientNet, **Tea-YOLO** | Accuracy: 85.2% |
| 20 | Pig facial expression recognition (Nie et al., 2023) | Faster RCNN, YOLO-(v4, v5s, v7, v8), **ASPP-YOLOv5** | Accuracy: 93.2% |
| 21 | Multi-class cotton weed detection (Dang et al., 2023) | YOLOv3 (tiny, SPP), **YOLOv4-pacasp** (s, x), Scaled-YOLOv4 (P5, 6, 7), YOLOR (P-6, CSP, CSP-X), YOLOv5 (n, s, m, l, x), YOLOv6 (n, s, m), YOLOv7(X, E6, D6) | Accuracy: 95.2% |
| 22 | Apple crop pest detection in a trap (Tian et al., 2023) | SSD512, Faster RCNN, YOLO-(v5m, v7m, v8m), RCNN, **MD-YOLO** | Accuracy: 86.2% |
| 23 | Maize tassels detection with UAV (Zhang et al., 2023) | CenterNet, Faster R-CNN, YOLO-(X, v4), **SwinT-YOLO (YOLOv4)** | Accuracy: 95.1% |
| 24 | Apple inflorescences detection (Xia et al., 2023) | SSD, FCOS, Faster RCNN, YOLO-(v4, v5s, X, v7), **MT-YOLOX** | AP: 83.4% |
| 25 | Unopen cotton bolls detection (Liu et al., 2023) | Faster RCNN, SSD, RetinaNet, YOLO-(v3, v4, v5, X), **MRF-YOLOX** | AP: 92.7% |
| 26 | Poultry chicken detection (Chen et al., 2023) | YOLOv7, YOLOv7x, **YOLOv7-tiny** | Accuracy: 98.2% |

| 27 | Apple detection with depth data (Kumar & Kumar, 2023) | Faster RCNN, YOLO- (v3, v5, v7), **Imp-YOLOv7** | Accuracy: 91% |
| 28 | Estrus cow detection (Wang et al., 2024) | SSD, Faster RCNN, **E-YOLO**, YOLO-(v5n, v5s, v7-tiny, v8n, v8s) | Accuracy: ~ 92% |
| 29 | Melon leaf disease detection (Xu et al., 2022) | RetinaNet, Faster RCNN, YOLO-(v3, v4, v5), **Prunned-YOLOv5s** | Accuracy: 96.7% |
| 30 | Crop disease and pest detection (Qing et al., 2023) | Faster RCNN, YOLO-(v4, v4-tiny, v5s, v5n), **YOLOPC (YOLOv5-n)** | Accuracy: 94.5% |

*Footnote: \*best performing model is highlighted in bold text and Ap = Average precision.*

Table 3: Data acquisition and processing method used in SLR documents.

| # | Objection or recognition task | Image acquisition | | # Class | Data Augmentation operation | | Input image resolution |
|---|---|---|---|---|---|---|---|
| | | Resolution | # images | | Transforms | #Images | |
| 1 | Broiler stun state (Ye et al., 2020) | 240×320 | 2319 | 3 | Flip, rotate | 27828 (x12) | 224×224 |
| 2 | Sugar beet damage (Nasirahmadi et al., 2021) | 1024×768 | 3425 | 1 | Flip, rotate | - | 608×608 |
| 3 | Orange (Mirhaji et al., 2021) | 3000×4000/ 2592×3872 | 766 | 1 | Rotate, contrast, shadow, sun flare, blur | 1568 (x2) | 416×416 |
| 4 | Fish uneaten pellet (X. Hu et al., 2021) | 1920×1080 | 175 | 1 | Contrast, mosaic | - | 416×416 |
| 5 | Fish behavior (Hu et al., 2021) | 4000×3000 | 2806 | 6 | Flip, rotate, blur | 72000 (x25.6) | 416×416 |
| 6 | Apple fruit (Chen et al., 2021) | 1632×1232 | 2950 | 1 | Rotate, saturation, hue, hist eq., mosaic | 10100 (x3.4) | 416×416 |
| 7 | Apple flower (Chen et al., 2022) | 1280×720/ 3024×3024 | 1800 | 6 | Flip, cutout, scale, brightness | 4000 (x2.2) | 640×640 |
| 8 | Date palm trees (Jintasuttisak et al., 2022) | 5472×3648 | 125 | 1 | Scale, crop, rotate, and color adjust. | 625 (x5) | 608×608 |
| 9 | Strawberry stage (An et al., 2022) | 1920×1080 | 5600 | 5 | Flip, scale, noise, rotate, mixup, mosaic | 28000 (x5) | - |
| 10 | Apple fruit (Zhang et al., 2022) | - | 1800 | 2 | Rotate, pan, flip, mosaic | 4860 (x2.7) | 416×416 |
| 11 | Maize weed (Liu et al., 2022) | 3024×4032 | 1000 | 5 | Rotate | 2400 (x2.4) | 608×608 |
| 12 | Flower (Park & Park, 2023) | 1920×1080 | 117 | 3 | Rotate, flip | 11808 (x100) | 608×608 |
| 13 | Grape stem (Wu et al., 2023) | - | 300 | 1 | Brightness, flip, noise | 1470 (x5) | 640×640 |
| 14 | Pig facial expression (Nie et al., 2023) | 720×480 | 2500 | 2 | Contrast, brightness, flip, rotate | 5000 (x2) | 608×608 |
| 15 | Cotton weed (Dang et al., 2023) | - | 5648 | 12 | Flip, rotate, blur, contrast, noise | 11296(×2) /22592(×4) | 640×640 |
| 16 | Apple crop pest (Tian et al., 2023) | 4288×2848 | 231 | 3 | Rotate, flip, brightness, scale | 631 (x2.7) | 512×512 |
| 17 | Apple inflorescences (Xia et al., 2023) | 4032×3024/ 2592×1944 | 524 | 1 | Mosaic | - | 1440×1088 |
| 18 | Unopen cotton bolls (Liu et al., 2023) | 1080×1440 | 750 | 1 | Noise, brightness, rotate, flip | 1500 (x2) | 640×640 |
| 19 | Poultry chicken (Chen et al., 2023) | 1920×1080 | 1000 | | Hue, saturation, brightness, flip, mosaic, mixup | - | 416×416 |
| 20 | Apple detection (Kumar & Kumar, 2023) | 2688×1520 | 1014 | 1 | Brightness, flip, rotate, blur, noise | 15210 (x15) | - |

*Table 4: An overview of YOLO network modification for various agricultural tasks.*

| Detection task | Model (Network) | Modification/improvements | Performance improvement |
|---|---|---|---|
| Broiler stunned state (Ye et al., 2020) | YOLO+MRM (ResNet) | Multilayer residual module (MRM): to learn finer image features of different classes | Accuracy: ↑6.66% |
| Fish uneaten pellet (X. Hu et al., 2021) | Imp-YOLOv4 (CSPDarknet53) | (1) PANet: to extract fine-grained features of small fish (2) Residual network was replaced with densely connected network to boost training speed (3) Reduction in CNN layers and replacing residual blocks to dense connection blocks: to reduce computation | AP: ↑27.21% and computation ↓30% |
| Fish behavior (Hu et al., 2021) | Imp-YOLOv3-Lite (MobileNetv2) | (1) MobileNet2: lightweight network. (2) Improved Spatial Pyramid Pooling (SPP): to improve accuracy (3) GIoU loss: to reduce scale sensitivity | Speed: ↑220% and size: ↓94.4% |
| Apple fruit (Chen et al., 2021) | Des-YOLOv4 (DenseNet) | (1) DenseNet: lightweight network (2) AP-loss: to improve class loss (3) Soft-NMS: to solve the problem of missing detection under overlapping conditions | Accuracy: ↑3.2% |
| Apple flower (Chen et al., 2022) | Imp-YOLOv5s (DarkNet) | (1) Coordinate Attention Mechanism (AM): to improve feature extraction (2) additional detection layer: for small objects (3) Bi-directional feature pyramid network (BiFPN) feature fusion module: to optimize network | Accuracy: ↑1.7% and AP small object: ↑ 4.8% |
| Wheat spikes (Zhao et al., 2022) | OSWSDet (DarkNet) | (1) Circle smooth label: for angular prediction/orientation (2) Additional detection layer: for the tiny target (wheat spike), (3) CIoU loss: to define spatial relationship of the oriented box, (4) Rotation-NMS: for duplicate oriented boxes | Accuracy: ↑32.1% |
| Strawberry growth (An et al., 2022) | SDNet (-) | (1) C3HB block (backbone): to improve feature extraction (2) Normalization AM: to improve small target accuracy (3) SIoU loss: to improve accuracy | Accuracy: ↑4.08% |
| Tomato (Ge et al., 2022) | YOLO-deepsort (ShuffleNet) | (1) ShuffleNetv2 (backbone): for lightweight (2) Convolutional block AM (CBAM): to extract fine features (3) BiFPN module: to improve accuracy (4) CIoU loss | Accuracy: ↑7.1% |
| Apple fruit (Zhang et al., 2022) | Imp-YOLOv4 (GhostNet) | (1) GhostNet: for lightweight (2) DW separable conv: to reconstruct neck and head (3) Coordinate AM: to improve feature extraction of small targets | Accuracy: ↑3.45%, speed: ↑5.7 fps, size: ↓84.46% |
| Seedling maize weed (Liu et al., 2022) | YOLOv4-weeds (DarkNet) | (1) Resblock and SE-block structure: to improve performance (2) Dense-SPP structure: to obtain rich features (3) Spatial AM: to improve accuracy (4) CIoU loss | Accuracy: ↑1%, speed: ↑8 fps, and size: ↓85.5% |
| Flower (Park & Park, 2023) | YOLOv4-circle (DarkNet) | (1) Circular bounding box and cIoU evaluation index: to improve accuracy | Accuracy: ↑2.26% |
| Crop row (rice) (Ruan et al., 2023) | Imp-YOLOR (GhostNet) | (1) GhostNet (backbone): for lightweight model (2) DW separable convolution: to reduce parameters (3) SPP and FPN channel AM: to strengthen feature extraction and fusion (4) Focal loss: to control weights of positive and negative samples | Accuracy: ↑2.20% and speed: ↑17.54fps |
| Cow behavior (Zheng & Qin, 2023) | PrunedYOLO-Tracker (DarkNet) | (1) Channel pruning: to reduce model complexity and size by eliminating unimportant channels (lightweight network). | Accuracy: ↑0.3%, size: ↓73.5%, param: ↓74.0% |
| Tea bud (Li et al., 2023) | Tea-YOLO (GhostNet) | (1) GhostNet (backbone): for lightweight model (2) DW separable convolution: to reduce computation and complexity (3) CBAM: to improve feature extraction (4) SIoU loss: to improve speed and accuracy | Accuracy: ↑1.08%, computation: ↓89.11%, param: ↓ 82.36% |

| Pig facial expression (Nie et al., 2023) | ASPP-YOLOv5 (DarkNet) | (1) Coordinated AM: to improve feature extraction and fusion (2) Atrous SPP module: to improve multi-scale feature to improve accuracy | Accuracy: ↑7 % |
|---|---|---|---|
| Apple crop pest (Tian et al., 2023) | MD-YOLO (DenseNet) | (1) DenseNet (backbone): to improve feature extraction, (2) Adaptive AM: to extract small target features, (3) Feature fusion module: to integrate spatial and semantic feature maps to improve accuracy, (4) CIoU loss: to optimize network | Accuracy: ↑7.8% |
| Maize tassels (Zhang et al., 2023) | SwinT-YOLO (Swin-transformer) | (1) Swin-Transformer (backbone): to improve global feature extraction (2) Modified PANet with DW separable convolution: to reduce computation | Accuracy: ↑2%. |
| Apple inflorescences (Xia et al., 2023) | MTYOLOX (Darknet +Swin-Transformer) | (1) DAT-Darknet (backbone): Deformable attention transformer layer to extract unique features (2) ST-PAFPN module: Swin-transformer based layer to aggregate multi-scale features to improve accuracy (3) GIoU loss: improve detection accuracy | Accuracy: ↑1.21% and speed: ↑1.8fps (limited improvement) |
| Unopen cotton bolls (Liu et al., 2023) | MRF-YOLO (DarkNet) | (1) Multi-receptive field extraction module (backbone): to enlarge network receptive field and reduce small feature loss (2) AM (bottle2neck_se multi-scale block): to enhance extraction of small cotton boll features (3) Additional detection layer: to improve accuracy of small targets | Accuracy: ↑7%. |
| Apple detection (Kumar & Kumar, 2023) | Imp-YOLOv7 (DarkNet) | Multi-head AM: to improve accuracy and depth estimation from RGB images | Accuracy: ↑5% |
| Estrus cow (Wang et al., 2024) | E-YOLO (Darknet) | (1) Triplet AM: to enhance object focus (2) Context information augmentation module: to enhance contextual information (3) Normalized Wasserstein Distance loss: to reduce sensitivity to position deviation of small target | Accuracy: ↑ 5.40% |
| Melon leaf disease (Xu et al., 2022) | Prunned-YOLOv5s (ShuffleNet) | (1) ShuffleNetv2 inverted residual block: to improve feature extraction (2) Channel Pruning: to reduce parameters (lightweight) | Accuracy: ↑6.2% and size: ↓85% |
| Crop disease & pest (Qing et al., 2023) | YOLOPC (-) | (1) PP-LCNet backbone: for lightweight model (2) Bottleneck with C2ff module: to improve performance (3) CBAM: to extract fine features | Accuracy: ↑4.7% |

Footnote: ↑ suggest improvement on model performance (speed and accuracy), while ↓ suggest reduction in model computation or size

The acquired images are auto-resized to a fixed resolution by YOLO, which can vary based on architecture and configuration. It takes square images as input, as shown in Table 3. At the same time, advanced versions can handle arbitrary image resolution provided both height and width are a multiple of 32 since YOLO down-samples input images at a scale of 32× (Bazame et al., 2021). The input image resolution significantly influences the YOLO detection performance (e.g., accuracy and speed). Higher-resolution images contain detailed information/features. Thus, a significant performance improvement was observed when increasing the image resolution from 416×416 to 896×896 pixels for coffee fruit detection (Bazame et al., 2021). The highest detection accuracy was observed at 800×800 resolution since YOLO layers (convolutional and pooling) gradually decrease the spatial dimension with increased network depth, making it difficult to detect small or overlapped objects (Bazame et al., 2021). Smaller objects typically carry smaller pixel footprints, which may disappear in the deeper network. Hence, higher image resolution is preferred. For example, an image resolution of 1400×1088 was selected for tiny, dense, and occluded apple inflorescence detection, which delivered accuracy of 83.4% (Xia et al., 2023).

Aerial images are captured at higher resolution and resized to the appropriate model input size. This typically impacts small and dense objects, which are prone to information loss during training. Therefore, aerial images of wheat fields captured at 5472×3648 resolution were cropped to 150×150 pixels to preserve or highlight small, dense wheat spikes while improving image processing and training efficiency (Zhao et al., 2022). Overall, image resolution influences the model performance, and detection accuracy tends to increase with higher resolution, but detection speed decreases for given computing resources. Subsequently, the processed images are split into a model development (train and validation) and evaluation (test) subset with more than 70% of data used for development. Test subset serves as independent data for unbiased model evaluation on unseen data and is considered optional but preferred in studies with large amounts of data.

### 4.2.3. Model Development and Modification

Deep learning requires large, quality, and diverse image data for model training, which is challenging and often impractical to collect. Thus, transfer learning is a widely accepted approach that involves using a pre-trained model, trained on a large dataset, and fine-tuned for new tasks with a smaller dataset (Yu et al., 2022; Zhuang et al., 2020). Transfer learning is inspired by humans' ability to transfer previously learned knowledge to solve similar new tasks by leveraging gained knowledge. The steps involved in transfer learning are presented in Figure 8. Transfer learning is the initial starting point and archives a good performance with reduced data and computational resources (Yu et al., 2022). Hence, all SLRs studied (n =30) implemented the transfer learning approach.

The YOLO pretrain models are trained on popular and large datasets, including COCO (Lin et al., 2015) and ImageNet (Deng et al., 2009), which include common objects. In transfer learning, pre-trained models are used as starting points and fine-tuned for new tasks with limited datasets. This model provides a strong foundation for generic object detection but often presents challenges when dealing with complex or specialized (agricultural) objects. Agricultural objects are complex and unstructured, which exhibit varying physical characteristics (e.g., size, shape, color, appearance) with growth stage, lightning, or changing environmental conditions. Moreover, pre-train models are not trained on complex objects with very fine and slightly less distinguishable features. For example, apple inflorescence, agricultural disease symptoms, insect and pest objects that are small, dense, and often occluded with fine features where pre-trained YOLO model accuracy largely suffered (Chen et al., 2022; Xia et al., 2023; Xu et al., 2022; Qing et al., 2023). The performance of generic object detection algorithms in agriculture primarily suffers for several reasons, including limited representation of complex objects, fine details and context, different object scales, and domain-specific challenges. Therefore, YOLO models are improved or modified to address a target-specific challenge, presented in Table 4 and discussed subsequently. The modifications often aim to improve the model performance regarding accuracy, speed, size, and robustness. Modifying network architecture (including backbone, neck, and head), loss functions, or training strategies are often implemented to achieve better detection performance.

Computer vision is critical to automated and robotic systems (e.g., real-time monitoring and advanced machinery). Therefore, the recognition accuracy and speed influence automated or robotic system's working capacity, efficiency, and effectiveness. A small increment in detection performance (accuracy or speed) will significantly influence the efficiency and scale of operation. For example, in robotic harvester, a 1% improvement in detection accuracy would increase yield by 1%. Table 4 discusses the performance improvement (i.e., accuracy, speed, and size) and the modification strategies adopted for specific object detection tasks. A (near) real-time object detection is crucial for multiple tasks, such as monitoring crops, animals (i.e., livestock, poultry, fish), and automated operation or handling, including fruit-vegetable sorting, harvesting, etc. The standard CNN network requires considerable storage and computational capacity and power that presents a challenge for deploying on resource-constrained devices (i.e., mobile, edge, single board, etc.), which are predominately used in agriculture to keep costs low (Zhang et al., 2022; Ruan et al., 2023). In lightweight networks, learnable parameters are reduced to achieve fast inference speed with reduced computational memory (size) while maintaining or improving accuracy. Several techniques can be employed for lightweight networks, which are discussed subsequently.

Lightweight structures networks: YOLO mainly employs "DarkNet" as the backbone for complex feature extraction, and several studies that employed the DarkNet have shown good detection performance (Table 4). DarkNet exploits multiple CNN layers for diverse and multi-scale object recognition with good accuracy, but relatively complex backbone networks that significantly influence the detection speed. Therefore, several authors designed different CNNs and structures to reduce parameters. MobileNet is an efficient CNN architecture suitable for low-memory devices (i.e., mobile and embedded devices). It reduces the parameters by integrating inverted residual structure and depth-wise (DW) separable convolution (Howard et al., 2017). The YOLOv3-Lite with MobileNetv2 was developed for real-time fish behavior monitoring in polyculture, reducing the model size from 401 MB to 22.5 MB and offering a detection speed of 240 fps with no reduction in accuracy (Hu et al., 2021). ShuffleNet is another computationally efficient CNN that uses pointwise group convolutions and channel shuffle operations to reduce computation costs (Ma et al., 2018; Zhang et al., 2017). A lightweight YOLOv5 with shuffleNetv2 was developed to detect greenhouse tomatoes at different growth stages, which reduced the model size to 10.5 MB (Ge et al., 2022). Likewise, GhostNet focus on efficiency by introducing a novel Ghost block, which aims to generate more feature maps with fewer parameters (Han et al., 2020; Tang et al., 2022). Automated agricultural machinery/robots require lightweight models with higher speeds or fps. Hence, an improved YOLOR with GhostNet was designed to detect rice row crops, improving speed by 114.1% (28.9 fps) while reducing accuracy by 2.20% (Ruan et al., 2023). A lightweight detection network was proposed for a tea-picking robot with real-time tea bud detection based on YOLOv4 and GhostNet. The improved "Tea-YOLO" network significantly reduced network parameters by 82.9% while reducing accuracy by 1.54% (Li et al., 2023). Similarly, an improved YOLOv4 with GhostNet was developed to detect apples for robotic picking, which improved speed by 5.7fps (45.2fps) with a slight improvement in accuracy, i.e., 3.45% (Zhang et al., 2022). Lightweight networks offer a trade-off between speed and accuracy. Hence, the network selection needs to be task-specific, and the authors often preferred faster models in exchange for slightly lower accuracy, where faster speed is a priority.

Model Compression techniques, such as pruning, quantization, and knowledge distillation, are employed to reduce network size with the least influence on accuracy (Li et al., 2023). Pruning is a widely used method to remove unnecessary/redundant parameters (i.e., connections, neurons, layers, channels, etc.) from the deep CNN, resulting in a sparser model with fewer parameters (Choudhary et al., 2020; Li et al., 2023). The recent YOLO (v5, v7, v8) networks are available with multiple scaled versions based on model complexity, which include nano (n), small (s), medium (m), large (l), and extra-large (x). The lighter version (e.g., n, s, m) has fewer parameters and offers faster speed at slightly reduced accuracy (Liu et al., 2022). However, they offer poor performance in general agricultural objects such as cows.

Meanwhile, heavier versions (e.g., l, x) provide good detection performance but are highly computational and slower in speed. Therefore, YOLOv5l was employed for multi-cow behavior monitoring and a channel pruning method was adopted to eliminate unimported channels, reducing the model size by 73.5% and slightly improving detection accuracy (Zheng & Qin, 2023). The pruned-YOLO delivered real-time cow behavior monitoring with a detection speed of 81fps. Likewise, a YOLOv5 with shuffleNetv2 was proposed for real-time melon leaf disease detection (Xu et al., 2022). The resulting lightweight network of 13.6 MB was further reduced to 1.1 MB (i.e., 85% reduction) with the channel pruning. It was deployed on an edge device (i.e., Jetson Nano), which delivered an inference speed of 13.8 ms (72.5fps) with 95.7% accuracy (Xu et al., 2022).

Convolutional is the building block of CNN, and the standard convolution kernel slides across the entire input image, performing computation at each position. Therefore, replacing standard convolutions with other convolutions (e.g., depth-wise separable and dilated) can reduce the parameters and computations, leading to a lightweight model (Zhang et al., 2022). The depth-wise (DW) separable convolution separates spatial and depth operations by using a separate kernel for each input channel (depth) with subsequent pointwise convolution operation (Guo et al., 2019). The DW separable convolutions provide an efficient alternative to standard convolutions and have been employed in several studies (Table 4), including Imp-YOLOv4 for apple fruit detection (Zhang et al., 2022), Imp-YOLOR for rice crop row detection (Ruan et al., 2023) and Tea-YOLO for tea bud detection (Li et al., 2023).

YOLO uses a DarkNet as a backbone for feature extraction and is commonly used in several studies (Table 4) along with other lightweight networks. Besides that, the DenseNet backbone was used for feature representation of apple crop pests (Tian et al., 2023) and apple fruit (Chen et al., 2021). DenseNet uses dense blocks to connect all layers with each other, forming a dense connection between layers (Huang et al., 2016). The dense connection allows efficient information flow and preserves fine-grained information during down-sampling operations to enhance feature extraction (Tian et al., 2023). Moreover, over the years, research has proposed several advanced features that can be integrated into YOLO to improve specific object detection tasks accuracy, speed, and robustness. The commonly used advanced features or techniques are discussed subsequently.

Attention module: The standard YOLO networks have a simple convolution structure for feature extraction, which often limits their ability to capture complex, diverse, dense, and small agricultural objects (Zhang et al., 2023). Therefore, the attention mechanism has shown promising results in enhancing feature extraction abilities through selective attention (Nie et al., 2023). Attention Mechanisms are inspired by the human vision system, which diverts attention to the essential regions in complex scenes. Likewise, in computer vision, the attention module focus on the most important image regions while disregarding irrelevant regions during training to enhance feature extraction, aiming to improve accuracy (Guo et al., 2022). Based on the operation domain, attention mechanisms can be categorized into: (i) channel, (ii) spatial, (iii) temporal, (iv) branch, (v) combined channel and spatial, and (vi) spatial and temporal. The channel attention focus on a channel domain to determine essential channels. For example, Channel attention was employed to pay attention to critical channels to detect rice row crops (Ruan et al., 2023) with YOLOR. While spatial attention focus on spatial domains to determine important spatial regions. Thus, it was used to improve the accuracy and robustness of "YOLOweeds" to detect seedling maize weeds (Liu et al., 2022). The convolutional block attention module (CBAM) combines channel and spatial attention in stacked series to enhance important channels and regions (Guo et al., 2022). CBAM's ability to leverage channel and spatial information enables capturing complex agricultural objects. Thus, it has been used in several studies, including tomato growth stage detection (Ge et al., 2022) to preserve the feature extraction ability of lightweight network (ShuffleNet), tiny tea bud detection in complex environments (Li et al., 2023) and to extract fine feature of citrus crop disease and pest (Qing et al., 2023). CBAM pays attention to individual domain information and lacks in capturing cross-domain information. Hence, a triplet attention mechanism was proposed for cross-domain interaction and was used to enhance the estrus cow features in "E-YOLO" (Wang et al., 2024). Likewise, CBAM often fails to capture long-range dependencies; hence, a coordinate attention module was proposed to integrate positional information into channel attention to focus on important regions with a larger receptive field (Guo et al., 2022). The following studies employed a coordinate attention mechanism: (1) to detect small apple flower buds in complex backgrounds, including morphological resemblance with fruits, leaves, and branches (Chen et al., 2022). (2) to improve the feature extraction of small and medium-sized apple fruits in complex backgrounds (Zhang et al., 2022). (3) to enhance the feature extraction ability of YOLO detecting pig facial expressions (Nie et al., 2023). The other variant of attention mechanism combined with YOLO to enhance feature extraction includes adaptive attention mechanism used for small pest detection in trap images (Tian et al., 2023), multi-head attention mechanism combined with YOLOv7 to distinguish similar apple objects in drone imagery (Kumar & Kumar, 2023) and bottle2neck_se multi-scale block in YOLOX for small cotton boll features extraction from large amount of image information (Liu et al., 2023). These studies report that YOLO integration with the attention mechanism delivered a better detection performance than standard YOLO.

Advanced feature extraction backbone: The CNN-based backbone network uses a stacked convolutional layer where the kernel works around the image pixels to extract specific features, which limits their ability to extract global features (Zhang et al., 2023) and model long-range representation (Guo et al., 2022). The vision transformer (ViT) has recently gained popularity in computer vision. Transformers replace traditional convolutional layers with self-attention mechanisms, allowing the model to capture long-range dependencies in images and the ability to extract global features (Khan et al., 2022; Xu et al., 2022). Therefore, an improved YOLOv4 was proposed with a Swin-transformer as the backbone with a multi-head attention mechanism to capture the global features of densely distributed maize tassels in drone imagery (Zhang et al., 2023). Likewise, a transformer module (i.e., DAT-Darknet, ST-PAFPN) based on multiple self-attention mechanisms is embedded into the YOLOX backbone and neck to extract global features of apple inflorescence (Xia et al., 2023). As shown in Table 4, both studies reported transformer-based models outperforming standard YOLO.

YOLO neck is responsible for backbone-extracted feature aggregation, refinement, and fusion. It aims to enhance the spatial and semantic details of multi-scale features. It includes convolutional layers, Spatial Pyramid Pooling (SPP), Feature Pyramid Networks (FPN), and other modules (Terven et al., 2023). Firstly, the SPP network can handle arbitrary input image size and is designed to capture multiscale feature information within an image. The SPP obtains rich and fine-grained image features to improve accuracy. Different SPP modules were implemented based on the detection task complexity. For example, an improved SPP for fish behavior (Hu et al., 2021), Dense-SPP for seedling maize weed detection (Liu et al., 2022), and Atrous SPP for pig facial expression (Nie et al., 2023). Secondly, the FPN detects objects at multiple scales with a top-down or bottom-up approach. Based on FPN, the Path aggregation network (PANet) improves the FPN and extracts find-grained features of small fish (Hu et al., 2021). The extension of FPN is a bi-directional feature pyramid network (BiFPN), which introduces bi-directional connections (top-down and bottom-up) between different feature pyramid levels to help information flow and integration (Ge et al., 2022). The BiFPN was used in the apple flower (Chen et al., 2022) and tomato (Ge et al., 2022) detection tasks to enhance further feature representation.

Loss function: The head makes a final prediction from the obtained feature map from the neck. The standard YOLO loss function is a sum of localization, classification, and confidence loss. During training, the loss function aims to optimize model weight to improve detection performance without affecting speed. Therefore, various loss functions have been proposed, which are discussed subsequently (Table 4). Localization loss is based on the Intersection over Union (IoU) score, which computes the degree of overlap between the predicted and ground truth bounding box. Since IoU only considers overlapped boxes (i.e., no overlaps result in zero IoU score) and has limitations while dealing with small, dense, or elongated objects. Therefore, an IoU extension called Generalized IoU (GIoU) and Complete IoU (CIoU) were introduced to address these limitations. GIoU modifies the IoU equation by adding a term that considers the smallest enclosed box area that completely encloses predicted and ground truth boxes (Rezatofighi et al., 2019). The GIoU loss was used in underwater fish (Hu et al., 2021) and apple inflorescence detection tasks (Xia et al., 2023). Further, CIoU extends the GIoU equation by adding terms to penalize differences in bounding box center distance, aspect ratio, and diagonal distance (Du et al., 2021). CIoU considers multiple aspects of predicted and ground truth boxes to measure bounding box prediction accuracy comprehensively. The CIoU loss was employed in various tasks, including wheat spikes detection (Zhao et al., 2022) to define spatial relationships, apple crop pest detection to optimize MD-YOLO during training (Tian et al., 2023), seedling maize weed (Liu et al., 2022), and tomato (Ge et al., 2022) detection task. Likewise, the SIoU loss function introduces directionality to bounding box regression, was used for detecting strawberry fruits closer to the size and location (An et al., 2022), and added a direction indicator in the tea bud detection model, leading to improved speed and accuracy (Li et al., 2023). Wang et al., (2024) proposed a new loss function called Normalized Wasserstein Distance loss to reduce sensitivity to position deviation of small targets in estrus cow detection task. Next, classification loss predicts accurate class labels, and cross-entropy is frequently used for multi-class classification. Besides this, average precision loss was used to solve the class imbalance problem in apple fruit detection (Chen et al., 2021), and focal loss was used in YOLO-R to improve classification scores for detecting rice crop rows (Ruan et al., 2023).

Additional detection layer in head: A standard YOLO includes three detection layers that downsample image dimension by 32, 16, and 8 to detect large, medium, and small size objects (Bochkovskiy et al., 2020; Redmon & Farhadi, 2016, 2018). These layers perform well for generic object sizes and often poorly on small or tiny-sized objects (Zhao et al., 2022). Therefore, to detect small objects such as wheat spikes (Zhao et al., 2022), apple flowers (Chen et al., 2022), and unopen cotton bolls (Liu et al., 2023), an additional detection layer was added to the network. This additional layer helped improve the detection accuracy of small-size objects (Table 4).

Other task-specific modifications include bounding box modification. YOLO outputs a horizontal-rectangular box since it works best for locating moving objects, which often may not adapt to varied shaped and multi-oriented agricultural objects, further influencing detection performance. For example, the generalized shape of fruits and flowers is circular, but using a rectangular box includes unimportant background regions, which may affect the ability to learn the YOLO model. Thus, a circular box was proposed for detecting circular chrysanthemum flowers, which contained more precise flower characteristics than a rectangular box, improving detection performance by 2.26% (Park & Park, 2023). Likewise, the box orientation is essential to improve detection performance; hence, to detect overlapped and dense wheat spikes in aerial imagery, an oriented bounding box was obtained with a smooth circle label module, which predicted the bounding box angle and location coordinates.

### 4.2.4. YOLO Integration

YOLO demonstrates excellent feature extraction abilities that can be further integrated with other state-of-the-art algorithms for specific applications. The YOLO output and bounding box can be directly counted in images for counting purposes. The counting data can be a crucial input for precision agricultural management and decision-making, including yield forecasting in row crops. For example, Zhang et al. (2023) counted the predicted maize tassel boxes in aerial images to predict maize yield. Liu et al. (2023) implemented YOLOX to detect and count unopened cotton bolls with 92% counting accuracy. Likewise, the counting data can be used to map the density of specialty crops. For example, Xia et al. (2023) mapped the detected apple inflorescence with a grid to understand the density of individual trees, which can be used for crop load management, including thinning. The orchard orange detection and counting via drone imagery was integrated with ArcGIS software to forecast orchard yield with less than 10% variation, which can be used for precision orchard management (Mirhaji et al., 2021). During the mechanical coffee fruit harvesting, geo-tagged images with real-time YOLO prediction capabilities were used for coffee fruit quality mapping in row crops to understand the spatial variability and future management planning with quality maps (Bazame et al., 2021). This approach works well in fixed images but often fails in moving frames or real-time mode, which counts duplicate objects. Therefore, YOLO can be integrated with object tracking algorithms such as SORT-Simple Online Real-time Tracking algorithms, which assign a unique ID to each object to avoid identity switches (Bewley et al., 2016). The YOLO output is an input to SORT, which uses Kalman Filters and Hungarian Algorithms to track objects. An integrated model combining YOLOv7+SORT+ autoregressive integrated moving average models was developed for automatic chicken monitoring systems in large poultry farms to alert potential health risks and hazards (Chen et al., 2023). DeepSORT is a SORT extension that introduces deep learning by adding appearance descriptions to the algorithm (Wojke et al., 2017). YOLO+DeepSORT was used to track and count tomatoes at different growth periods to predict yield in the greenhouse (Ge et al., 2022). Likewise, YOLOv5 was integrated with a multi-object tracker (Cascaded-Buffered IoU) to monitor dairy cow behavior (Zheng & Qin, 2023).

Automatic agricultural navigation requires expensive sensors, including (GPS, LiDAR, etc.)(Badgujar et al., 2023). Therefore, visual navigation was proposed with a simple camera system and models by integrating YOLOR with a density-based clustering algorithm and least squares method to extract the rice row crop line for autonomous machinery navigation (Ruan et al., 2023). The integrated framework precisely detected and extracted the rice row crops with accuracy above 90%. Likewise, depth estimation requires a specialized sensor such as LiDAR. Therefore, YOLOv7 with a multi-head attention mechanism was proposed to detect apple and depth estimation from drone RGB imagery (Kumar & Kumar, 2023). The object detection algorithm can accurately detect and locate agricultural objects (i.e., fruits and vegetables), an essential component of an automated robotic or harvesting system. However, stem localization or identifying fruit and vegetable picking or cutting spots is still challenging. Therefore, the object detection (YOLO) and key-point detection models (HRNet) were integrated to locate the grape stems for machine picking, which accurately detected the precise stem location with a detection accuracy of 92% (Wu et al., 2023).

YOLO exhibits excellent visual understanding capabilities, while recent large language models (GPT-4) provide deep logical reasoning capabilities. Thus, GPT-4 was integrated with YOLO to analyze pests and diseases in crops and generate a diagnostic report for the farmers. The YOLO predictions were converted into simple text and then fed to a GPT-4 to diagnose citrus disease and pests and provide explanations and suggestions (Qing et al., 2023). The proposed system demonstrated 90% reasoning accuracy in generating agricultural diagnostic reports and serving as a virtual pathologist or agronomist. These integrations further illustrate the applicability and adoption of YOLO for digital agricultural operations.

### 4.2.5. YOLO Strength and Limitations

Over the years, the YOLO algorithm has undergone multiple versions and improvements, each offering unique strengths and limitations on specific tasks and data characteristics. Mainly, agricultural operation requires accurate, fast, real-time, and robust object detection. YOLO generally fulfills these requirements with a certain degree of trade-off between speed and accuracy. YOLO models are fast, lightweight, and offer real-time detection performance with reasonable accuracy. For example, (1) YOLO+MRM detected more than 180,000 broilers/hours in an automatic broiler-slaughter line (Ye et al., 2020), (2) YOLOv4 delivered 29 fps with accuracy above 90% for mechanical sugar beet damage detection during harvester (Nasirahmadi et al., 2021). (3) YOLOv3-lite with 22.5MB size delivered 240fps for fish detection behavior (Hu et al., 2021). (4) YOLOv4 delivered 31.1 fps for the detection of rice row crops used for agricultural machinery navigation (Ruan et al., 2023). (5) Improved YOLO model delivered 81 fps for cow behavior monitoring task with good detection accuracy (Zheng & Qin, 2023). The lightweight YOLO is best suited for low-cost and resource-constraint embedded devices. Hence, several studies have demonstrated the ability of real-time detection with a low-cost edge device (Chen et al., 2021; Hu et al., 2021; Ruan et al., 2023).

Agricultural object variability and environmental complexity affects detection performance. Hence, researchers often emphasize collecting and training the model on large and diverse datasets at various growth, lighting, and multiple field locations or conditions to develop accurate and robust models. Diverse training images should be collected to address the environmental variability (such as lighting conditions, seasons, rainy, cloudy, bright, etc.). For example, orchard orange images were captured in different illumination conditions (i.e., cloudy, sunny, and night) to train robust YOLOv4 for detecting oranges (Mirhaji et al., 2021). The YOLOv4 performed well under different illumination conditions with no significant difference in night and daytime image predictions. Likewise, to detect apple flowers, images were collected at all blooming levels under different light, flower density, occlusion conditions, and interannual data (Chen et al., 2022). The trained YOLOv5 showed robustness to environmental factors and performed well on previous year's data, showing an interannual knowledge transfer capability. However, low light intensity at nightfall presented challenges for apple flower feature extraction, affecting model performance. Besides this, YOLOv4 showed robustness in detecting weeds in maize fields under different weather conditions (Liu 2022) and YOLOv3 showed robust performance against illumination and water quality variation in fish detection (Hu et al., 2021). Also, machine factors such as vibration may influence detection performance, but YOLOv3 used to detect coffee fruits showed robustness against vibration and varying fruit color during coffee harvesting (Bazame et al., 2021).

Deep learning models are inherently data dependent and require sufficient labeled image datasets, which are labor-intensive, time-consuming, costly, and require expertise. For example, 146,516 chicken objects were annotated in 1000 images to train YOLOv7 for automatic chicken monitoring in large-scale poultry farms (Chen et al., 2023). Meanwhile, Zheng & Qin, (2023) manually annotated 34,013 targets (cows) to train the cow behavior recognition system. In some instances, it often becomes challenging to capture a sufficient dataset; for example, an insect detection system would require images of multiple insect species, which would be challenging to obtain (Tian et al., 2023; Badgujar et al., 2023). In such cases, artificial data could be generated from limited datasets with data argumentation, computer graphics, or other methods. Besides this, YOLO training is a highly computational task. It often involves using parallel processing units such as a Graphical Processing Unit (GPU) to accelerate computation and reduce training duration. However, GPUs are expensive and require significant capital expenditure. All the studies in SLR reported the use of the best available computing resources to train models. To avoid expensive hardware (GPU), a few authors often preferred subscription-based cloud services called "google Collaboratory" (Mirhaji et al., 2021).

The standard YOLO often struggles to detect small and dense objects. Hence, several studies modified the YOLO architecture (i.e., backbone, neck, head) to improve feature extraction with an attention mechanism, additional detection layers, and other methods, which would require an interdisciplinary expertise and knowledge background. The well-trained YOLO models show a certain robustness to object variation. However, if the object domain changes, the models might need to be retrained with a new dataset. For example, the pig facial expression detection model may need to be retrained if the pig breeds change (Nie et al., 2023). Likewise, the fruit detection model may need retraining if the fruit variety has been upgraded (Mirhaji et al., 2021; Park & Park, 2023; Zhang et al., 2022).

## 5. Conclusions

In conclusion, using the YOLO algorithm for agricultural object recognition is a big step forward for digital agricultural tools and technology. Utilizing YOLO's real-time, multi-object identification capabilities, a reliable system has been developed to recognize and locate various agricultural objects, including crops, pests, diseases, animals and so on. Once trained on large images, YOLO algorithms are capable of recognition the small and fine objects which may often requires an expertise. This review presents the current scientific landscape of YOLO in agriculture along with research trends, hotspot and unexplored research domain in broad agriculture. The systematic literature review provides a comprehensive overview on YOLO model development in agriculture from data collecting to deployment. This study also identifies the limitations of standard YOLO model to detect complex agricultural object often found in dynamic environments. We also discussed the potential YOLO network modification techniques aiming to improve either accuracy, speed, robustness and reduce computation. We provide research evidence which shows the effectiveness of the YOLO-based agricultural object recognition system and its adaptability to various illumination settings, environmental conditions, and object sizes. The algorithm's single-pass image processing capability has shown to be extremely useful in the dynamic and resource-intensive environment of large-scale agricultural landscapes. Furthermore, the YOLO technique in agricultural object identification can be used for more than only object detection; it can also be used for essential aspects of precision agriculture like pest control, anomaly detection, and crop yield estimation. Farmers and other agricultural professionals may optimize resource allocation and reduce environmental impact using YOLO to make real-time informed decisions. As we move forward, agricultural workflows that incorporate YOLO-based solutions have the potential to completely transform how the sector approaches overseeing and controlling large-scale farming operations. The results of this study add to the growing body of knowledge in computer vision applications in agriculture by providing a practical and expandable framework for resolving issues arising from contemporary farming methods. In summary, the YOLO algorithm is a potential approach that supports effective, sustainable, and technologically advanced farming operations when it comes to agricultural object detection. This study lays the groundwork for additional investigation and application, representing a significant advancement in incorporating state-of-the-art computer vision technology into the agricultural environment.

**Declaration of Competing Interest**

There is no conflict of interest to describe regarding this manuscript.

**Acknowledgement CRediT**

Chetan Badgujar: Conceptualization, Methodology, Data collection, Data curation, Writing- Original draft preparation. Alwin P.: Visualization, Investigation, Supervision, Writing- Original draft preparation. Hao Gan.: Validation, Supervision, Reviewing, and Editing.